%% file: main.tex
\documentclass{article}
\usepackage[final]{corl_2022}

\usepackage{amsmath}
\usepackage{amsfonts}
\usepackage{amsthm}
\usepackage{svg}
\usepackage{subfigure}
\usepackage{booktabs}
\usepackage{tikz}
\usetikzlibrary{shapes.geometric}
\usetikzlibrary{shapes,backgrounds}
\usepackage{algorithm}
\usepackage{algpseudocode}
\usepackage{capt-of}
\usepackage{varwidth}
\usepackage[font={footnotesize}]{caption}
\usepackage{wrapfig}

\newtheorem{theorem}{Theorem}
\newtheorem{definition}{Definition}

\newtheorem{remark}{Remark}

\usepackage{titlesec}

\title{Task-Relevant Failure Detection for Trajectory Predictors in Autonomous Vehicles}

\author{
  Alec Farid\thanks{Equal Contribution} \hspace{1.5pt}$^{,}$\thanks{Work done as an intern with the NVIDIA Research Autonomous Vehicle Research Group}\hspace{3pt}$^{,1}$, Sushant Veer$^{*,2}$, Boris Ivanovic$^2$, Karen Leung$^{2,3}$, Marco Pavone$^{2,4}$\\
  $^1$Princeton University,
  $^2$NVIDIA Research, $^3$University of Washington, $^4$Stanford University\\
  \texttt{afarid@princeton.edu, \{sveer,bivanovic,kaleung,mpavone\}@nvidia.com} \\
}

\begin{document}
\maketitle
\setcounter{footnote}{0} 

\vspace{-2em}
\begin{abstract}
In modern autonomy stacks, prediction modules are paramount to planning motions in the presence of other mobile agents. However, failures in prediction modules can mislead the downstream planner into making unsafe decisions. Indeed, the high uncertainty inherent to the task of trajectory forecasting ensures that such mispredictions occur frequently. Motivated by the need to improve safety of autonomous vehicles without compromising on their performance, we develop a probabilistic run-time monitor that detects when a ``harmful" prediction failure occurs, i.e., a \emph{task-relevant} failure detector. We achieve this by propagating trajectory prediction errors to the planning cost to reason about their impact on the AV. Furthermore, our detector comes equipped with performance measures on the false-positive and the false-negative rate and allows for data-free calibration. In our experiments we compared our detector with various others and found that our detector has the highest area under the receiver operator characteristic curve. 
\end{abstract}
\keywords{Run-time Monitoring, Autonomous Vehicles, Trajectory Prediction}

\input{sections/introduction}
\input{sections/related_work}
\input{sections/preliminaries}

\input{sections/failure_detection}
\input{sections/experimental_results}
\input{sections/limitations-future-conclusion}

\clearpage
\acknowledgments{The authors are grateful to Anirudha Majumdar for helpful feedback and discussions on this work.}

\bibliography{main.bbl}

\newpage
\input{sections/appendix}

\end{document}

%% file: sections/introduction.tex
\section{Introduction}
\label{sec:intro}

In general, autonomous vehicle (AV) stacks forecast the trajectories that non-ego agents---such as other vehicles, pedestrians, bicyclists, etc.---are likely to take. The accuracy of these predictions is crucial for the motion planner to ensure that the AV does not find itself in an unsafe state. Unfortunately, trajectory forecasting is an endeavor wrought with a high degree of uncertainty, a large portion of which, arguably, is aleatoric, arising from the stochasticity of human behavior. The situation is further exacerbated by the epistemic uncertainty that arises from unavoidable learning errors when training trajectory-prediction networks from finite driving logs. Therefore, it should come as no surprise that the trajectory-prediction networks frequently mispredict, leading to misguided motion plans. How then, in the face of such uncertainty, do we ensure safety of the AV without compromising its functionality? We approach this challenge by developing a \emph{task-relevant} failure detector for the prediction module that only triggers if the estimate of the realized cost at a given time-instant is significantly worse than the cost predictions based on trajectory forecasting.

\begin{figure}[t]
% \vspace{-1.4em}
  \centering
  \subfigure[Task-relevant prediction failure]
    {
    \includegraphics[width=0.48\textwidth]{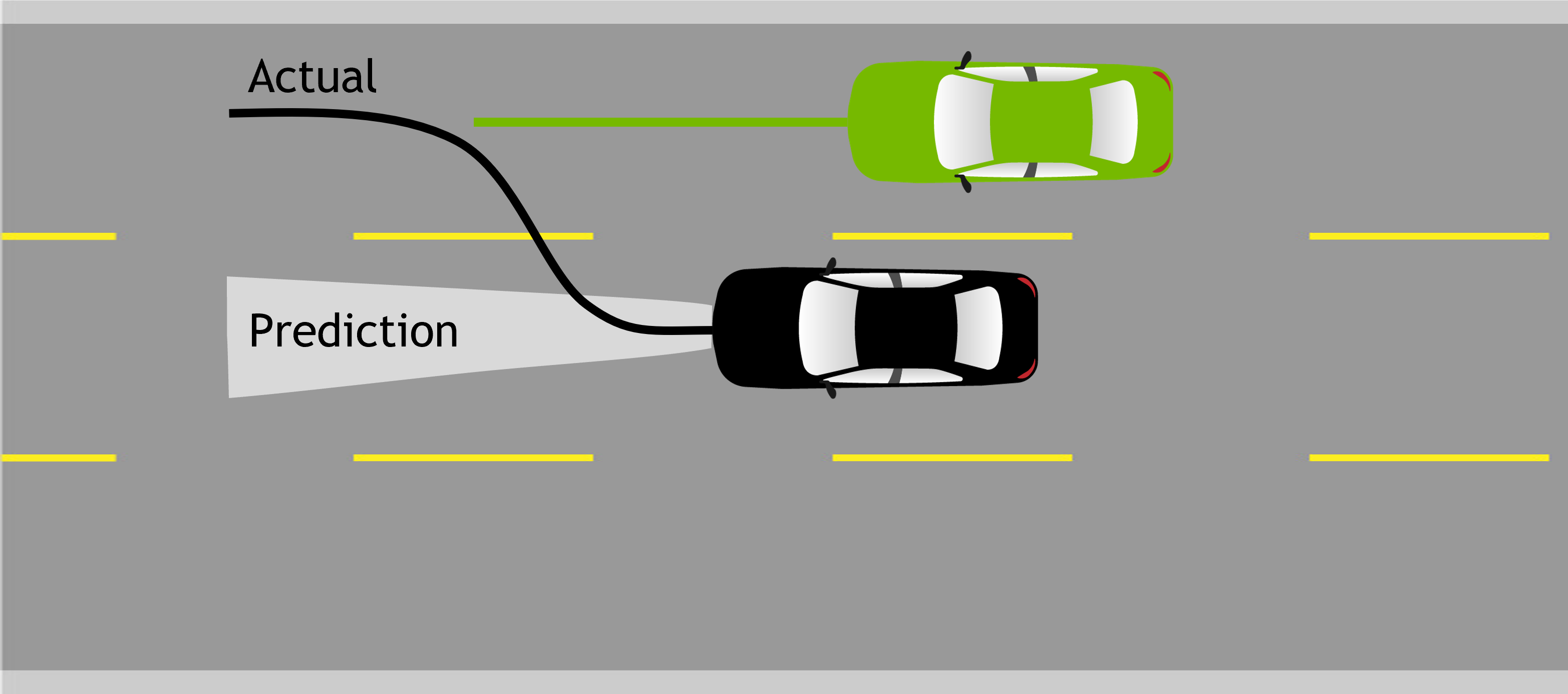}
    \label{fig:task-relevant}
    }
  \subfigure[Non-task-relevant prediction failure]
    {
    \includegraphics[width=0.48\textwidth]{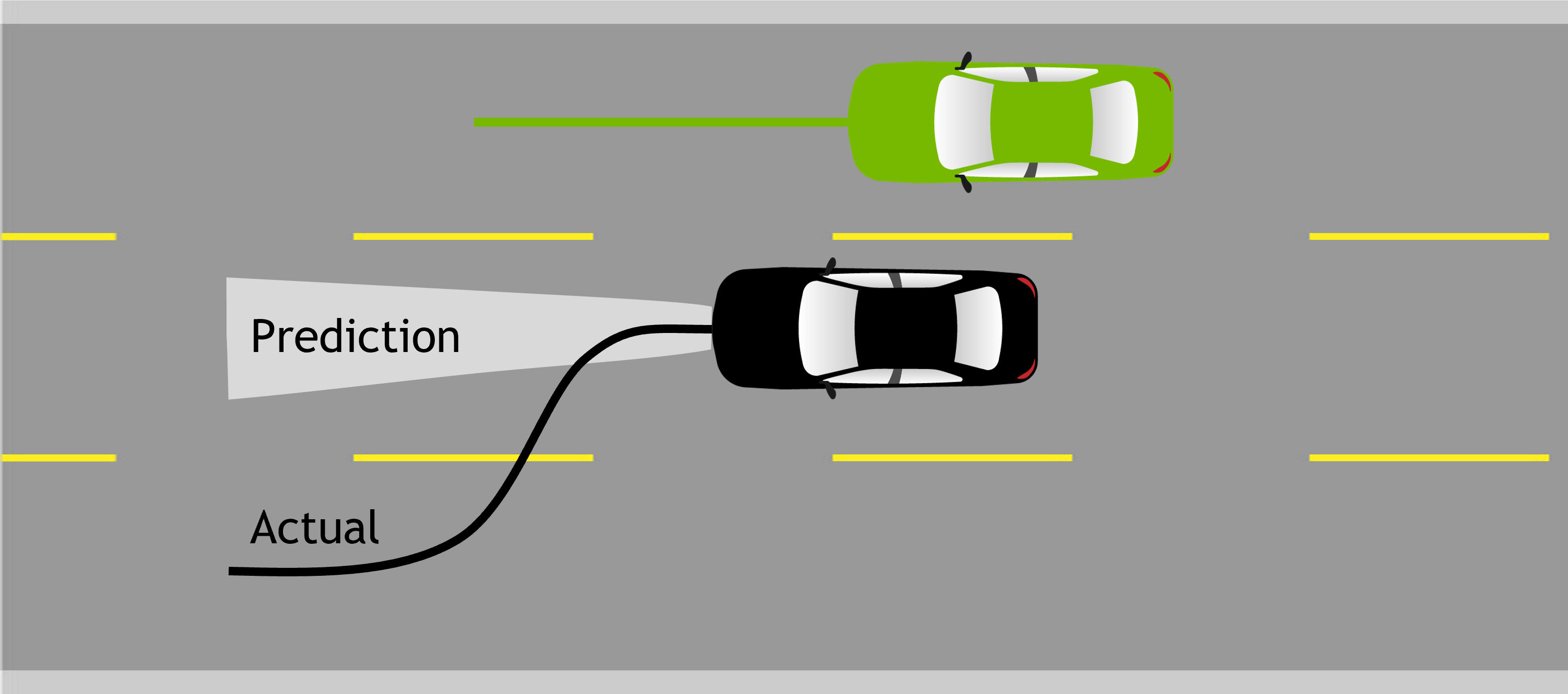}
    \label{fig:not-task-relevant}
    }
  \caption{\footnotesize Illustration of task-relevant failure detection. The ego-vehicle and its motion plan is in green, the non-ego vehicle and the trajectory it takes is in black while the predicted trajectory distribution is in light-grey. \label{fig:anchor}}
  \vspace{-5mm}
\end{figure}

We require our detector to embody the following desiderata: First, the detector should be \emph{task relevant} to reduce the number of unnecessary interventions that hurt the AV's functionality, i.e., the detector should only trigger if the failure is harmful to the AV's safety. To better illustrate task-relevance, consider the scenarios in Fig.~\ref{fig:anchor}. If the non-ego vehicle is predicted to stay in its lane, but it turns right, into the AV's lane (Fig.~\ref{fig:task-relevant}), the cost for the planner would increase and therefore, the task-relevant detector should trigger. However, if the car turns left, away from the AV's lane (Fig.~\ref{fig:not-task-relevant}), the cost for the planner would decrease, hence, the detector should not trigger. Note that in both the scenarios in Fig.~\ref{fig:anchor}, the predictions are incorrect, but they are task-relevant only in Fig.~\ref{fig:task-relevant}. Second, the detector should be an independent module, i.e., it should not modify the prediction module that it monitors. This enables easier integration of the detector with existing AV stacks and also facilitates the prediction module to specialize in its task without having to forego performance in favor of detectability of failures. Finally, the detector should be accompanied with a \emph{metric to gauge its performance quality} to facilitate interpretable and trustworthy decision making. Most prior work satisfies only a subset of these desiderata \cite{LeungSchmerlingEtAl2019, Liu18, Farid21, Greenberg21}. In this paper we propose an approach that satisfies all these three desiderata by reasoning about the effect of erroneous predictions further downstream at the planner module of the AV stack.

\textbf{Statement of Contributions.} The main contributions of this paper are as follows: (i) We formalize a \emph{notion of anomaly} that underpins the design of a task-relevant prediction failure. (ii) We develop a \emph{fast and efficient task-relevant failure detector} for the prediction module used in AV stacks which is simultaneously accompanied by \emph{guarantees on the false-positive and false-negative detection rates} for improved interpretability. (iii) Owing to the interpretability of our approach, we can perform \emph{data-free calibration} for our detector. Furthermore, we show in our experiments that our data-free calibrated detector's performance is not very far from the optimally calibrated detector using the Receiver Operator Characteristic (ROC) curve. (iv) We demonstrate the efficacy of our approach on nuScenes \citep{nuscenes} and nuPlan teaser \citep{nuplan} datasets and provide comparisons with other detection methods in the literature. We find that our detector has the highest Area Under the ROC (AUROC) curve.

%% file: sections/related_work.tex
\section{Related Work}
\label{sec:related-work}

\textbf{AV Safety.} 
Since autonomous vehicles require interactions with other road users and are safety-critical by nature, there has been a lot of effort, from regulators, industry, and researchers alike, in ensuring safe AV operations \cite{ieee2022P2846,nister2019safety,shalev2017formal, althoff2014online,kianfar2013safety,GMSafety2018, LyftSafety2020, ZooxSafety2021, WaymoSafety2021, NVIDIASafety2021} % ArgoSafety2021,
(see \cite{dahl2019collision} for a review). A common approach is to compute ``inevitable'' collision sets (ICS), e.g., via Hamilton-Jacobi reachability computation \cite{mitchell2005time}, with selected assumptions on other agents' behaviors, and perform \textit{shielding} where the AV will flag a situation as unsafe whenever it is close to entering the ICS and execute an appropriate evasive action (e.g., \cite{LeungSchmerlingEtAl2019, hsu2022sim,Haimin22}). A primary challenge is selecting \textit{reasonable} behavior assumptions for ICS computation to balance tractability, interpretability, and compatibility with real-world driving interactions \cite{leung2021towards,tian2022safety}. Most of these approaches assume that other agents act in an adversarial manner leading to overly conservative ICS. Instead, in this work, we build a run-time monitor that identifies only those prediction failures which are safety-critical in nature. Furthermore, the general detection approach we present could also be used as a run-time task-relevant safety monitor.

\textbf{Anomaly Detection.} 
Anomaly detection or change detection in signal processing literature has been an active area of work for decades (see~\citep{Basseville88} for a review). The goal is to identify events which deviate significantly from a reference input. Such methods have been applied in supervised learning settings to higher-dimensional inputs such as images~\citep{Hendrycks17, Lee18, Hendrycks20} (see also~\citep{Ruff21} for a review); however, these methods are not task-relevant in nature. Other methods perform detection by estimating agent~\citep{Bera16, Wiederer22} or input~\citep{SharmaAzizanEtAl2021} atypicality. Recent techniques also allow for detection using the reward in reinforcement learning settings using streams of sensor inputs~\citep{Greenberg21, Sedlmeier19, Sedlmeier20, Cai20, Wu21}. However, these methods do not provide guarantees on the false-positive and false-negative detection rates. Some methods that can provide guarantees come with caveats such as specific architectures~\citep{Liu18,Siddiqui16}, only providing guarantees in retrospect (i.e. after multiple anomalies have occurred)~\citep{Farid21}, or on test data that is statistically similar to the training data~\citep{Farid22}. We develop an \textit{online} anomaly detector which provides guarantees for both false positive rate and false negative rate. Unlike many of the works described above, our detector does not require modifying the existing module on which it operates. 

\textbf{Planning-aware Prediction.} The vast majority of behavior prediction works focus solely on the task of prediction (improving forecasting accuracy, incorporating additional sources of scene context, accelerating runtime, etc), in isolation from other components of the autonomy stack~\citep{RudenkoPalmieriEtAl2019}. In recent years, however, there has been increasing interest in better integrating prediction with the rest of the robotic autonomy stack. Examples include works that more tightly integrate prediction with object classification~\citep{IvanovicLeeEtAl2021}, tracking~\citep{WengYuanEtAl2021,WengIvanovicEtAl2022b}, and planning~\citep{ZieglerBenderEtAl2014,LiuLeeEtAl2017,FanZhuEtAl2018,ZengLuoEtAl2019,IvanovicElhafsiEtAl2020,ChenRosoliaEtAl2020,SchaeferLeungEtAl2021}. While these approaches focus on better \emph{architectural} integrations across the stack, a few recent works have also proposed to integrate modules \emph{metrically}. For instance, \cite{IvanovicPavone2022,McAllisterWulfeEtAl2022} present metrics that evaluate perception and prediction modules in a planning-aware fashion, e.g., by heavily weighting detection or prediction errors that could cause a collision. Our work is similar in that it can identify anomalous predictions which would detrimentally affect ego-vehicle planning, however, it does so online with guaranteed false-positive and false-negative detection rates.

%% file: sections/preliminaries.tex
\section{Problem Definition}

We express the dynamics of the AV and the scene around it as a Partially-Observable Markov Decision Process (POMDP) defined as a tuple $(\mathcal{X}, \mathcal{O}, \mathcal{U}, f)$ where $\mathcal{X}$ is the state space which is composed of the state of the ego agent $x^{\rm e}$, non-ego agents $x^{\rm ne}$, and other variables such as the environment map $x^{\rm m}$, $\mathcal{O}$ is the space of observations that the ego agent receives from its sensors, $\mathcal{U}$ is the space of the control inputs for the ego agent, and $f(x_t|x_{t-1},u_{t-1})$ is the stochastic state-transition function. 

Let $c_t:\mathcal{X}\times\mathcal{U}\to[0,\infty)$ be a cost function at a time instant $t$. Our framework is agnostic to the choice of the cost function; we only assume that the cost function is formulated such that higher values imply worse / less safe outcomes for the AV. As an example cost function is the closest distance of an AV to other non-ego agents and obstacles in the map. Given the predicted state $\hat{x}_{t+\tau}$ and the control input $u_{t+\tau}$ at $\tau$ time steps in the future, the predicted cost is $\hat{c}_{t+\tau}:=c_{t+\tau}(\hat{x}_{t+\tau},u_{t+\tau})$.

With a slight abuse of notation, let $x_t \in \mathcal{X}$ be an \emph{estimate} of the state (instead of the true state) at time $t$ provided by the perception module by processing the sensor observation $o_t$. The prediction module takes in a history of state estimates $x_{0:t}$ and provides a distribution $\psi(\hat{x}^{\rm ne}_{t+1:t+T}|x_{0:t})$ on the predicted future state trajectories $\hat{x}^{\rm ne}_{t+1:t+T}$ of non-ego agents over the horizon $T$. The distribution produced by the prediction module induces a sequence of distributions $\{\phi_{t+\tau}(\hat{c}_{t+\tau}|x_{0:t})\}_{\tau=1}^T$ on the predicted cost $\hat{c}_{t+\tau}$ for each time-step in the prediction horizon $T$. The observed cost $c^\star_{t+\tau}:= c_{t+\tau}(x_{t+\tau},u_{t+\tau})$ at time-step $t+\tau$ is the cost estimated or ``observed'' by the AV given state-estimates of the ego and non-ego agents $x_{t+\tau}$ and the control input $u_{t+\tau}$. We define the notion of anomaly that will underpin our approach to prediction failure detection:
\begin{definition}[$p$-quantile anomaly]\label{def:anomaly}
For any time-step $t+\tau$, let $\phi_{t+\tau}(\hat{c}_{t+\tau}|x_{0:t})$ be the distribution on the predicted cost as described above. If the observed cost $c^\star_{t+\tau}$ lies in the top $p$-quantile of $\phi_{t+\tau}(\hat{c}_{t+\tau}|x_{0:t})$, then the prediction module suffers from a $p$-quantile anomaly. 
\end{definition}

\begin{figure}[b]
% \vspace{-1mm}
  \centering
    \includegraphics[width=0.80\textwidth]{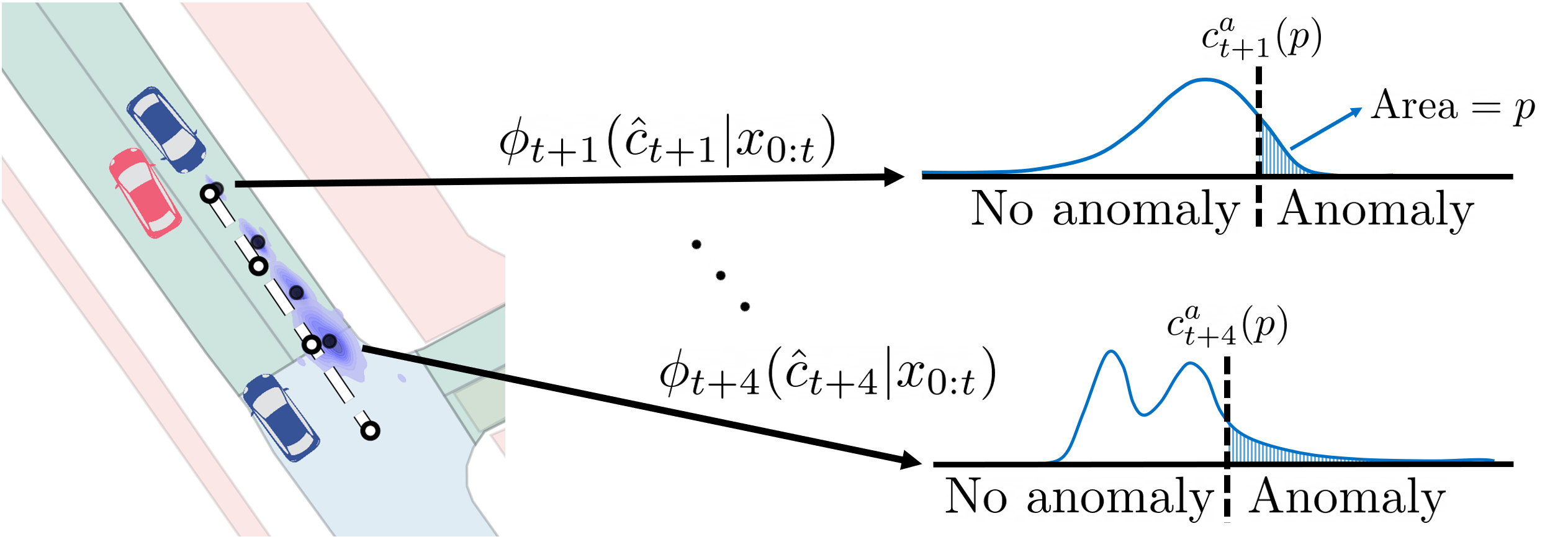}\vspace{4pt}
  \caption{\footnotesize Illustration of the concept of $p$-quantile anomalies. The ego vehicle is red, non-ego vehicles are blue, ground truth trajectories are white, and trajectory predictions are the purple contour plots. At each time step, the cost function induces a distribution $\phi_{t+\tau}$ on predicted costs depicted by the blue plots on the right. If the observed cost $c^\star_{t+\tau}$ is within the top-$p$ quantile of $\phi_{t+\tau}$, it is a $p$-quantile anomaly. \label{fig:quantile-anomaly-detection}}
\end{figure}

The observed cost $c^\star_{t+\tau}$ is, theoretically speaking, not necessarily drawn from the predicted cost distribution $\phi_{t+\tau}$ since the observed cost might depend on unmodeled/unobservable factors. However, under nominal circumstances (i.e., few or no task-relevant prediction errors and no major deviations by the AV from the open-loop motion plan), it is reasonable to expect the observed cost to be similar to the sampled predicted costs. This motivates the definition of anomaly as per Def.~\ref{def:anomaly} and the design of a prediction failure detector based on the concept of $p$-quantile anomalies (illustrated in Fig.~\ref{fig:quantile-anomaly-detection}). Importantly, the notion of $p$-quantile anomaly, being a cost-based quantity, intrinsically captures the notion of task-relevance. Note that if the predictor predicts cost-degrading behavior correctly, the observed cost would not lie in top $p$-quantile as this is in alignment with the predictions.

%% file: sections/failure_detection.tex
\section{$p$-Quantile Anomaly Detector}
\label{sec:task-relevant}

In this section, we present an algorithm for detecting $p$-quantile anomalies along with guaranteed false-positive and false-negative detection rates afforded by our algorithm. Let $c^{\rm a}_{t+\tau}(p)$ be the cost which separates the top $p$ proportion of $\phi_{t+\tau}(\hat{c}_{t+\tau}|x_{0:t})$ from the bottom $1-p$. If $c^\star_{t+\tau} \geq c^{\rm a}_{t+\tau}(p)$, then we have a $p$-quantile anomaly. The lack of an analytical expression\footnote{Even if the distribution on the predicted trajectories $\psi$ takes the form of a Gaussian mixture model \citep{salzmann20}, the nonlinear cost function may prevent the predicted cost distribution $\phi$ from acquiring an analytical form.} for $\phi$ prevents us from exactly computing $c^{\rm a}_{t+\tau}(p)$, hence, we use a sampling-based approach for detecting if the observed cost lies within the $p$-quantile of the distribution on the predicted cost; see Fig.~\ref{fig:quantile-anomaly-detection} for an illustration of the $p$-quantile anomaly within the trajectory prediction framework.

Suppose at time $t+\tau$ we sample $M$ predicted costs $S_{t+\tau} := \{\hat{c}^1_{t+\tau}, \dots, \hat{c}^M_{t+\tau}\}$ independent and identically distributed (i.i.d.) from $\phi_{t+\tau}(\hat{c}_{t+\tau}|x_{0:t})$ and arrange them in an increasing order (i.e., $\hat{c}_{t+\tau}^1\leq \hat{c}_{t+\tau}^2 \leq \cdots \leq \hat{c}_{t+\tau}^M$). Our detector $D$ returns a positive detection if the observed cost is greater than at least $M-n$ predicted costs in $S$, where $n\in\{0,\cdots,M-1\}$: 
\begin{align}\label{eq:detector-1}
    D(n, S) & := c^\star_{t+\tau} \geq \hat{c}_{t+\tau}^{M-n} \enspace .
\end{align}
Intuitively, by checking where the observed cost ranks against the sampled predicted costs, our detector estimates the quantile of $\phi$ in which the observed cost lies. In the rest of this section we present our detection algorithm, provide bounds on it's false-positive and false-negative detection rates, and use these bounds to analytically calibrate the detector by choosing appropriate $M$ and $n$.
% \vspace{10mm}

\subsection{Detection Algorithm}
The detector, Algorithm~\ref{alg:detector}, is executed at each planning cycle. We sample $M$ trajectories for each non-ego agent from the trajectory-prediction network $\psi$. Inference from prediction networks is usually a bottleneck in real-time control; however, note that for most AV stacks, we do not need to sample these trajectories specially for detection, as they would be drawn regardless for use in planning. Hence, by reusing these samples our detector does not generate significant computational overhead. At each planning cycle, our detector receives $M$ predicted costs for each non-ego agent at each prediction time-step (see line 1 in Algorithm~\ref{alg:detector}). For each subsequent time-step in the prediction horizon, we monitor $D$ in \eqref{eq:detector-1}. If the detection criteria is met, the detector returns \texttt{True} along with the identifier of the non-ego agent responsible for the detection. In Sec.~\ref{subsec:exp-task-relevant}, we show that this detector runs in less than $0.1$ milliseconds (ms) which is sufficiently fast for real-time monitoring.
% \vspace{-1mm}
\begin{algorithm}[h]
\caption{$p$-Quantile Anomaly Detector (QAD) \label{alg:detector}}
\small
\begin{algorithmic}[1]
	\State \textbf{Input:} For each non-ego agent, a sequence of sets $S_{t+\tau}=\{\hat{c}_{t+\tau}^1,\cdots,\hat{c}_{t+\tau}^M\}$ for each $\tau\in\{1,\cdots,T\}$ 
	\State \textbf{Input:} Cost functions $c_{t+\tau}$ for each $\tau\in\{1,\cdots,T\}$ 
	\For{wall-clock time $w$ in $t+1:t+T$}
	    \For{each non-ego agent}
	    \State $c_w^\star \leftarrow c_w(x_w,u_w)$ 
	    \If{$c_w^\star \geq \hat{c}_w^{M-n}$}
	        \State \Return (\texttt{True}, non-ego agent Id)
	    \EndIf
	    \EndFor
	\EndFor
	\end{algorithmic}
\normalsize
\end{algorithm}

\subsection{Bounds on False-positive and False-negative Rates}
\label{subsec:fp-fn}
Let us first define the false-positive rate (FPR) and the false-negative rate (FNR) for $D$.
\begin{definition}[False-positive Rate]\label{def:fpr-def}
FPR of the detector is defined as the probability of drawing $S_{t+\tau}$ such that a $p$-quantile anomaly detection occurs when the anomaly does not occur, i.e.,
\small
\begin{align}\label{eq:fpr-def}
    \hspace{-2mm} \text{FPR} := \underset{S_{t+\tau}\sim\phi_{t+\tau}^M}{\mathbb{P}}[D(n,S_{t+\tau}) \wedge c^\star_{t+\tau} < c^{\rm a}_{t+\tau}(p)] & = \underset{S_{t+\tau}\sim\phi_{t+\tau}^M}{\mathbb{P}}[c^\star_{t+\tau} \geq \hat{c}_{t+\tau}^{M-n} \wedge c^\star_{t+\tau} < c^{\rm a}_{t+\tau}(p)] \enspace.
\end{align}
\normalsize
\end{definition}

\begin{definition}[False-negative Rate]\label{def:fnr-def}
FNR of the detector is defined as the probability of drawing $S_{t+\tau}$ such that a $p$-quantile anomaly detection does not occur when the anomaly does occur, i.e.,
\small
\begin{align}\label{eq:fnr-def}
    \hspace{-2mm} \text{FNR} := \underset{S_{t+\tau}\sim\phi_{t+\tau}^M}{\mathbb{P}}[\neg D(n,S_{t+\tau}) \wedge c^\star_{t+\tau} \geq c^{\rm a}_{t+\tau}(p)] & = \underset{S_{t+\tau}\sim\phi_{t+\tau}^M}{\mathbb{P}}[c^\star_{t+\tau} < \hat{c}_{t+\tau}^{M-n} \wedge c^\star_{t+\tau} \geq c^{\rm a}_{t+\tau}(p)] \enspace.
\end{align}
\normalsize
\end{definition}

Ideally, for safety-critical systems, such as AVs, we desire a low FNR to ensure that we do not miss any unsafe scenarios. This can be trivially achieved by having a detector that triggers all the time (high FPR), but clearly that is not desireable for the AV's performance. Hence, a good detector must also exhibit a low FPR. Motivated by this, we provide bounds on the FPR and FNR for our detector.

\begin{theorem}[FPR and FNR Bound] \label{thm:fpr-fnr}
Let the detector $D$ for the occurrence of a $p$-quantile anomaly be defined as in \eqref{eq:detector-1}. Then, the FPR (Def.~\ref{def:fpr-def}) and the FNR (Def.~\ref{def:fnr-def}) for the detector satisfies:
\begin{align}\nonumber
    \text{FPR} \leq \sum_{i=0}^n \binom{M}{i} p^i (1-p)^{M - i}=:\overline{\text{FPR}} \enspace, & & \text{FNR} \leq \sum_{i=n+1}^M \binom{M}{i} p^i (1-p)^{M - i}=:\overline{\text{FNR}} \enspace.
\end{align}
\end{theorem}
The proof of this theorem is presented in App.~\ref{app:proof}. Thm.~\ref{thm:fpr-fnr} ensures that if the detector identifies an anomaly (i.e., $D$ holds true), then with probability at least $1-\overline{FPR}$, the observed cost does lie in the top-$p$ quantile of $\phi_{t+\tau}$, while if our detector does not detect an anomaly (i.e., $D$ is not true), then with probability at least $1-\overline{FNR}$, the observed cost does not lie in the top-$p$ quantile of $\phi_{t+\tau}$. 

Our detector satisfies all the three desiderata laid out in Sec.~\ref{sec:intro}. First, our detector encodes the notion of \emph{task-relevant failure detection} since it only triggers when the observed cost is high compared to the predicted costs, in the sense of the $p$-quantile anomaly (Def.~\ref{def:anomaly}). Second, the detector is \emph{independent of the trajectory prediction module} that it monitors. Finally, we provide bounds on FPR and FNR of our detector in Thm.~\ref{thm:fpr-fnr} that serve as a metric on the detector's performance. These bounds facilitate greater interpretability and promote easier detector calibration.

\vspace{-0.3mm}
\subsection{Data-free Detector Calibration}
\vspace{-0.4mm}
\label{subsec:calibration}

The analytical upper bounds for the FPR and the FNR in Thm.~\ref{thm:fpr-fnr} allows data-free calibration of $D$ by letting us choose $M$ and $n$ so as allow a user to specify FPR and FNR values. However, we can only provide a strong bound on either the FPR or the FNR, but not both. This follows from the observation that the upper bounds in Thm.~\ref{thm:fpr-fnr} satisfy $\overline{\textit{FPR}}+\overline{\textit{FNR}}=1$. It is worth clarifying that this is \emph{not} a fundamental limitation of our detector, but that of our analysis. We show in Sec.~\ref{sec:result} that our detector, in fact, achieves low FPR and low FNR simultaneously.

%% file: sections/experimental_results.tex
\vspace{-0.3mm}
\section{Experimental Results}
\vspace{-0.5mm}
\label{sec:result}

In this section, we benchmark the performance of our detector on nuScenes \citep{nuscenes} and nuPlan teaser \citep{nuplan} datasets against various other methods. We study the detector in two applications: (i) run-time monitoring of task-relevant prediction network failures and (ii) filtering interesting scenarios from unlabelled expert driving logs. Through these experiments we demonstrate the ability of our detector to provide low FPR and low FNR for detecting task-relevant mispredictions. All experiments were conducted on a desktop computer with an Intel i9-10980XE CPU (18 cores) and 64 GB of RAM. Our code is available at: \href{https://github.com/NVlabs/pred-fail-detector}{\texttt{https://github.com/NVlabs/pred-fail-detector}}.

\textbf{Trajectory prediction module.} While our method is agnostic to the particular prediction method used, in our experiments we employ Trajectron++~\citep{salzmann20}, a state-of-the-art multi-agent trajectory forecasting model, to predict the motion of non-ego agents. 
It is a graph-structured recurrent neural network that predicts an agent's future position distribution given its past trajectory history and the past trajectories of its neighboring agents.
At its core, Trajectron++ uses a conditional variational autoencoder (CVAE) to model the potential for multiple future trajectories.

\textbf{Detection and comparison methods.} We compare our detector---$p$-quantile anomaly detection (QAD)---against four other approaches: (i) \emph{likelihood detection} baseline which triggers when the likelihood of the achieved states of an agent in the predicted GMM is below a threshold; (ii) \emph{uniform and partial cost degradation} tests \citep{Greenberg21} which uses $p$-values to detect shifts in the cost distribution; (iii) \emph{time-to-collision (TTC)} detector which triggers when the TTC drops below a threshold; and (iv) detection based on \emph{Hamilton-Jacobi (HJ) reachability} analysis \cite{LeungSchmerlingEtAl2019} which triggers if a collision is possible assuming an adversarial behavior by the non-ego agent. See Sec.~\ref{app:comparisons} for more information about the comparison methods.

\subsection{Task-relevant Detection}
\label{subsec:exp-task-relevant}

\textbf{Overview.} We study the use of our detector and other baseline detectors mentioned above as run-time monitors for task-relevant prediction failure. The experiments are conducted in the recently released nuPlan teaser dataset \citep{nuplan} which allows the inclusion of reactive agents in the driving logs.

\textbf{Planner.} We assume a behavior generator provides us with a reference trajectory to follow. Our planner is then tasked with following this trajectory while avoiding collision with other agents on the road, ensuring ego comfort and satisfying the ego's dynamical constraints. The planner achieves this by generating a motion primitive tree \citep{SchmerlingLeungEtAl2018} and choosing the trajectory that minimizes the weighted sum of a series of cost functions; further details on the planner are supplied in App.~\ref{app:planning}. We use time-to-collision cost $c_{\text{ttc}}$ (computed assuming that agents continue moving at a constant speed in their current heading), momentum-shaped distance to other agents cost $c_{\text{d2a}}$, distance to goal cost $c_{\text{d2g}}$, distance to the reference trajectory cost $c_{\text{d2r}}$, speed limit cost $c_{\text{velocity}}$, ego comfort cost $c_{\text{comfort}}$, and reversing cost $c_{\text{reverse}}$ which penalizes the car for going against the heading direction of the lane:
\begin{align}
    c_{t+\tau} = w_1 c_{\text{ttc}} + w_2 c_{\text{d2a}} + w_3 c_{\text{d2g}} + w_4 c_\text{d2r} + w_5  c_{\text{velocity}} + w_6 c_{\text{comfort}} + w_7 c_{\text{reverse}} \enspace.
\end{align}
The exact expressions for the cost terms and the weights $w_1, \cdots, w_7$ used in the experiments are provided in App.~\ref{app:cost-func}. The motion planner uses trajectory forecasts of other non-ego agents drawn from Trajectron++ to compute the momentum-shaped distance and the time-to-collision cost. The cost functions take about $1 \pm 0.5$ ms to evaluate. Note that these cost evaluations for planning are reused by our anomaly detector in Algorithm~\ref{alg:detector}, thereby minimizing any excessive computational overhead by our detector. The anomaly detection computation takes less than 0.1 ms.

\textbf{Dataset and Labeling.} We test on 260 nuPlan \citep{nuplan} driving scenarios which comprises of a total of 2568 prediction horizons. We hand-label each prediction horizon for task-relevant prediction errors by visually inspecting plots of true trajectories of agents along with their predicted trajectories. We emphasize that the labeling is done merely by visual inspection and does not use the planning costs or prediction likelihoods. This is to ensure that the labeling procedure is independent of the detection methods to promote fair comparisons; see App.~\ref{app:hand_label} for more details on the labeling procedure.

\begin{remark}[Imbalance in Dataset]\label{rem:imbalance}
Despite the use of reactive agents along with our custom planner in nuPlan, there is a strong bias in the dataset towards benign scenarios. Only 3.5\% of this dataset is labeled as positive. Due to the imbalance, there is a bias toward detectors to exhibit lower FPR and higher FNR, since false-negative errors are amplified while the false-positive errors are mitigated. 
\end{remark}

\begin{table}[b]
\hspace{-15pt}
\begin{minipage}{0.43\textwidth}
  \centering
    \includegraphics[width=1\textwidth]{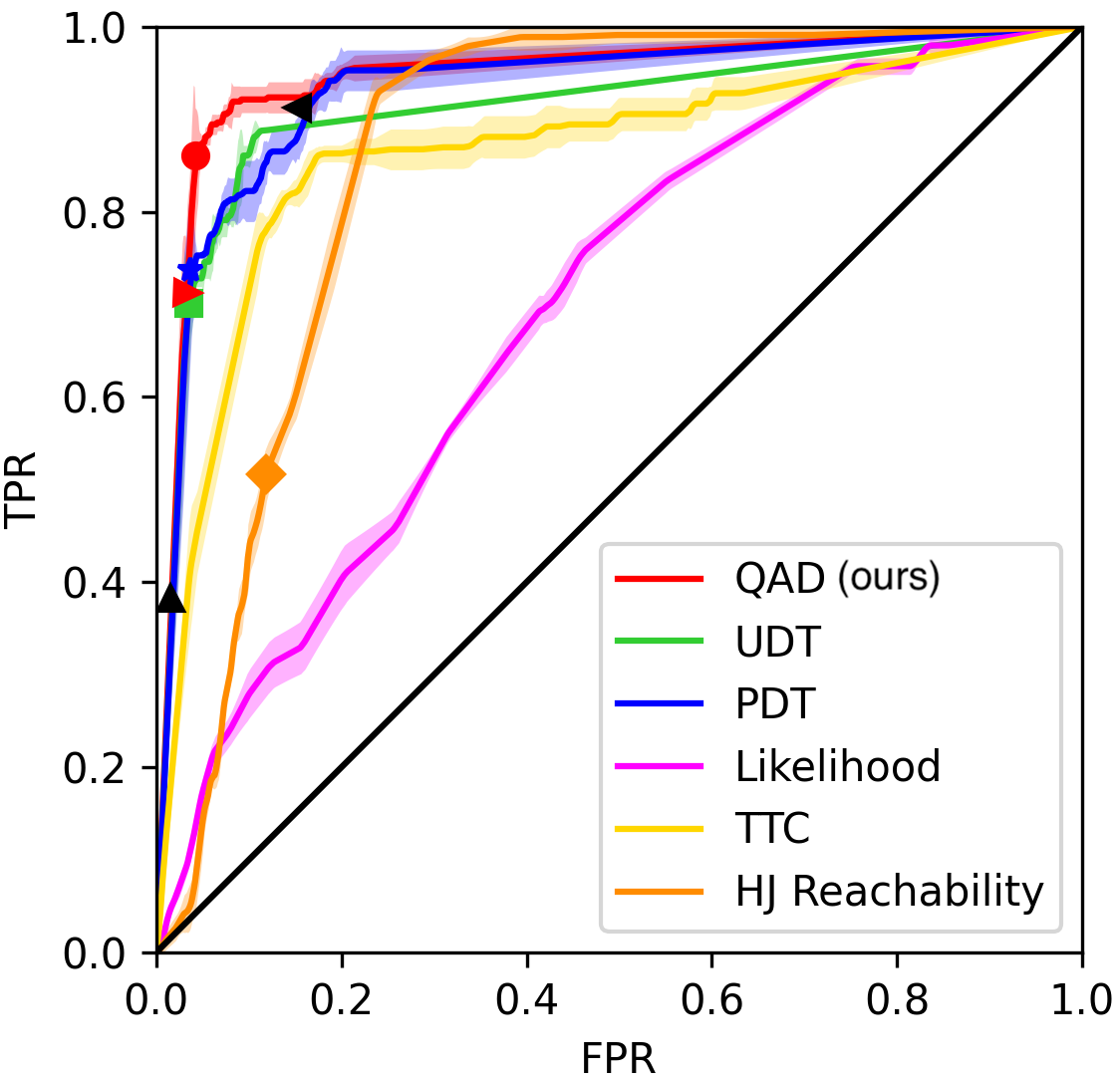}
\captionof{figure}{\footnotesize ROC curve for detectors. We report the mean of $5$ trials, the shaded region is $2$ times the standard deviation. \label{fig:roc_comparisons}}
\end{minipage}
\begin{minipage}{0.57\linewidth}
    \captionof{table}{AUROC and best point from ROC ($\uparrow$ AUROC) \label{table:best_point_on_roc}}
    \setlength{\tabcolsep}{2.5pt}
    \centering
    \footnotesize
    \begin{tabular}{cccc}
        \toprule
        Detection Method & AUROC & FPR \% & FNR \% \\
        \midrule
        QAD (ours) & \textbf{0.946$\pm$0.002} & \textbf{8.4$\pm$0.4} & 7.9$\pm 0.7$ \\
        PDT \citep{Greenberg21} & 0.932$\pm 0.005$ & 18.5$\pm 0.2$ & \textbf{5.8$\pm$1.1} \\ 
        UDT \citep{Greenberg21} & 0.912$\pm 0.002$ & 10.6$\pm 0.3$ & 11.5$\pm 0.4$ \\ 
        TTC & 0.864$\pm 0.004$ & 17.3$\pm 0.5$ & 13.7$\pm 0.4$ \\
        HJ Reachability \cite{LeungSchmerlingEtAl2019} & 0.862$\pm 0.003$ & 23.7$\pm 0.3$ & 7.2$\pm 0.6$ \\
        Likelihood & 0.693$\pm 0.006$ & 45.9$\pm 0.4$ & 24.5$\pm 0.4$\\
        \bottomrule
    \end{tabular}
    \normalsize
    \vspace{4pt}
    \captionof{table}{Data-free calibration ($\uparrow$ distance from $y=x$) \label{table:data_free_calibration}}
    % \vspace{-6pt}
    \setlength{\tabcolsep}{2.5pt}    
    \centering
    \footnotesize
    \begin{tabular}{ccccc}
        \toprule
        Detection Method & 
        & FPR \% & FNR \% & Dist. \\
        \midrule
        QAD (FNR) (ours) & \tikz{\draw[red,fill=red]  circle(0.5ex);} & 4.2$\pm 0.3$ & 13.9$\pm 1.1$ & \textbf{0.58$\pm$0.01} \\
        FNR Ensemble (ours) & \tikz{\node[regular polygon, draw = black, regular polygon sides = 3, fill = black, scale=0.35, rotate=90] (p) at (0,0) {};} & 15.2$\pm 0.3$ &  \textbf{8.7$\pm$1.2} & 0.54$\pm$0.01 \\
        PDT (FPR) \citep{Greenberg21}  & \tikz{\node[star, regular polygon sides = 3, fill = blue, scale=0.5] (p) at (0,0) {};}  & 3.7$\pm 0.3$ & 26.3$\pm 0.6$ & 0.50$\pm$0.01\\
        QAD (FPR) (ours) & \tikz{\node[regular polygon, draw = red, regular polygon sides = 3, fill = red, scale=0.35, rotate=270] (p) at (0,0) {};} & 3.4$\pm 0.3$ & 28.8$\pm 0.9$ & 0.48$\pm$0.01 \\
        UDT (FPR) \citep{Greenberg21} & \tikz{\node[regular polygon, draw = green, regular polygon sides = 4, fill = green, scale=0.5] (p) at (0,0) {};}  & 3.5$\pm 0.3$ & 29.9$\pm 0.9$ & 0.47$\pm$0.01 \\
        HJ Reachability \cite{LeungSchmerlingEtAl2019} & \tikz{\node[regular polygon, draw = orange, regular polygon sides = 4, fill = orange, scale=0.45, rotate=45] (p) at (0,0) {};} & 11.8$\pm 0.2$ & 48.6$\pm 1.7$ & 0.28$\pm$0.01\\
        FPR Ensemble (ours) & \tikz{\node[regular polygon, draw = black, regular polygon sides = 3, fill = black, scale=0.35, rotate=0] (p) at (0,0) {};} & \textbf{1.5$\pm$0.3} &  61.6$\pm 1.8$ & 0.26$\pm$0.01\\
        \bottomrule
    \end{tabular}
    \normalsize
\end{minipage}
\vspace{-1em}
\end{table}

\textbf{Results.} We plot the Receiver Operator Characteristic (ROC) curve in Fig.~\ref{fig:roc_comparisons} for each detector by varying: the detection rank $M-n$ in \eqref{eq:detector-1} for our detector, the $p$-value for UDT and PDT, the likelihood threshold for the likelihood-based detector, the TTC threshold for TTC detector, and the sub-level set of the value function for the HJ reachability based detector. The ROC curve for our detector is the closest to an ideal detector and has the highest Area Under the ROC (AUROC) curve. We obtain the best calibration for each detector by choosing the point that is farthest from the $y=x$ line in Fig.~\ref{fig:roc_comparisons} and report the empirical FPR and FNR for them in Table~\ref{table:best_point_on_roc}. Our detector has the lowest FPR and the third-lowest FNR. PDT has a slightly better FNR than our method, but the FPR is much higher, while for all other methods both FPR and FNR are higher than our method. As one would expect, the naive likelihood baseline does not perform very well while the HJ reachability detector is overly conservative exhibiting a lot of false-positives.

We also compare the performance of our detector with data-free calibration using Thm.~\ref{thm:fpr-fnr} with the data-free calibration of UDT, PDT, and HJ reachability. For our detector, we consider an anomaly to have occurred when the true cost lies in the top-$5\%$ of the predicted cost distribution (see Def.~\ref{def:anomaly}); we chose the $5\%$ quantile because it is a commonly used threshold for detection in statistics. We include two versions of our detector in this study: (i) calibrated to bound the FPR below $5\%$ and (ii) calibrated to bound the FNR below $5\%$ using Thm.~\ref{thm:fpr-fnr}; as discussed in Sec.~\ref{subsec:calibration}, we cannot simultaneously obtain a strong bound on both FPR and FNR. We use a $p$-value of $0.05$ for UDT and PDT, and a $0$-sublevel set of the value function for HJ reachability. Additionally, we construct an FPR ensemble detector, which only detects positive if \textit{all} of QAD (FPR), UDT, PDT, and HJ reachability are positive, and an FNR ensemble detector, which detects positive if \textit{any} of QAD (FNR), UDT, PDT, and HJ reachability are positive. For the ensemble detectors, we use the data-free calibration for each method described above.

These detectors are labeled using markers in the ROC curves in Fig.~\ref{fig:roc_comparisons}. Their empirical FPR and FNR is reported in Table~\ref{table:data_free_calibration} and we sort the table by distance from the line $y=x$.
We observe that our FPR bounded detector satisfies the $5\%$ bound provided by Thm.~\ref{thm:fpr-fnr}; however, our FNR bounded detector violates the bound. This disagreement between the theory and the empirical performance for FNR occurs because of the mismatch between our anomaly criteria of top-$5\%$ quantile and what the human data labeller considers to be an anomaly. A higher empirical FNR suggests that the anomalous quantile should be expanded beyond $5\%$; case in point, the optimal $p$-quantile anomaly detector obtained using the ROC curve in Table~\ref{table:best_point_on_roc} with $7\%$ FNR corresponds to approximately the top-$35\%$ quantile as an anomaly. Furthermore, the imbalance in the dataset also contributes to this problem, as discussed in Rem.~\ref{rem:imbalance}. More data as well as a more balanced dataset can perhaps remedy this mismatch and lower the FNR; we will explore this in future work. Despite this mismatch, we see in Table~\ref{table:data_free_calibration} that our FNR detector performs the best out of all detectors in terms of distance to the $y = x$ line in the ROC plot. The FNR ensemble and PDT also perform well, favoring false-negatives and false-positive respectively. Since false-negatives can be safety critical, a designer may choose our FNR ensemble detector at the cost of more false-positives. Ultimately, our FNR detector is not very far off from the optimal detector obtained from the ROC curve, i.e., with \textit{no data at all} we are able to calibrate our detector using Thm.~\ref{thm:fpr-fnr} to perform very well.

\subsection{Filtering expert driving logs}
\textbf{Overview.} Given driving data logs, we consider the task of filtering scenarios with anomalies that may be of interest for robsutifying the prediction network or for diagnostic evaluations. We use the nuScenes and nuPlan datasets as the driving logs. For the likelihood detector we choose the threshold $0.05$, and for TTC we choose 1 second, while for all other detectors, we use the data-free calibration specified in Sec.~\ref{subsec:exp-task-relevant}.

\textbf{Dataset.} The dataset consists of 375 nuPlan and 152 nuScenes scenarios with a total of 3890 prediction horizons, in all of which the ego-vehicle is driven by a human driver and there are no collisions.

\textbf{Cost functions.} Since the nuScenes and nuPlan driving logs are collected by a human driver, we do not have access to the true planning ``cost" used by the driver. Instead, we use a weighted sum of cost terms contributed by the momentum-shaped distance and the time-to-collision for each non-ego agent as a proxy cost: $c_{t+\tau} = w_1 c_{\text{ttc}} + w_2 c_{\text{d2a}}$. The exact expressions of $c_{\rm ttc}$ and $c_{\text{d2a}}$ are provided in App.~\ref{app:cost-func}. Our results suggest that this cost function is well-aligned with human intuition as it allows us to filter out interesting scenarios while ignoring the uninteresting ones.

\begin{figure}[t]
\vspace{-1em}
  \centering
    \subfigure[]{
    \includegraphics[width=0.23\textwidth]{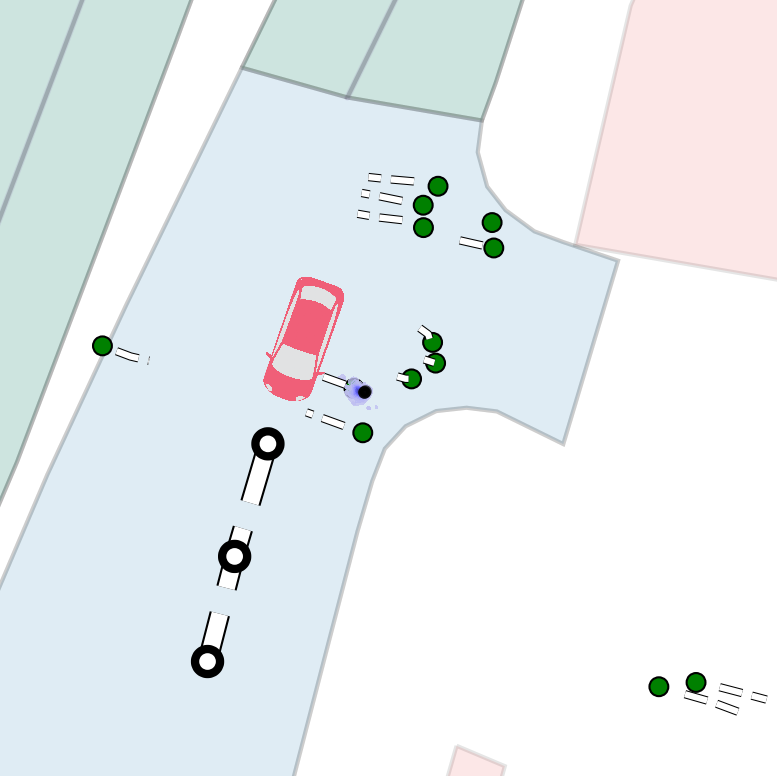}
    }
    \subfigure[]{
    \includegraphics[width=0.23\textwidth]{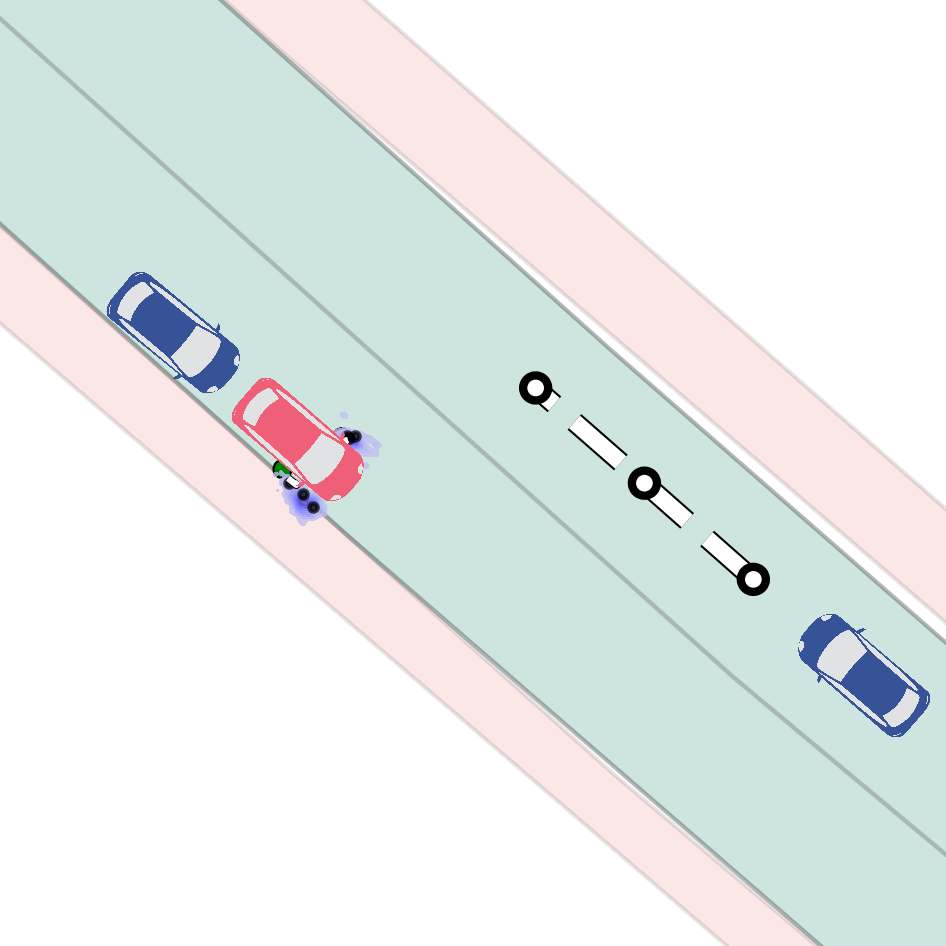}
    }
    \subfigure[]{
    \includegraphics[width=0.23\textwidth]{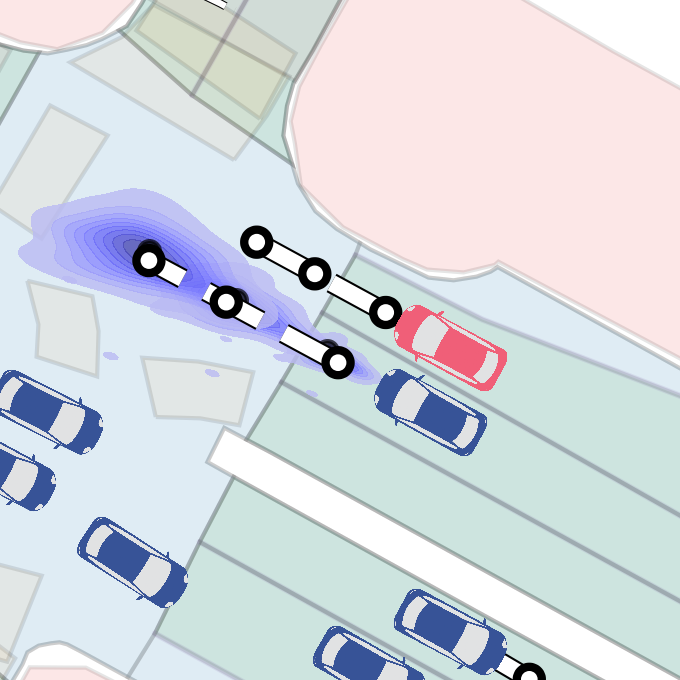}
    }
    \subfigure[]{
    \includegraphics[width=0.23\textwidth]{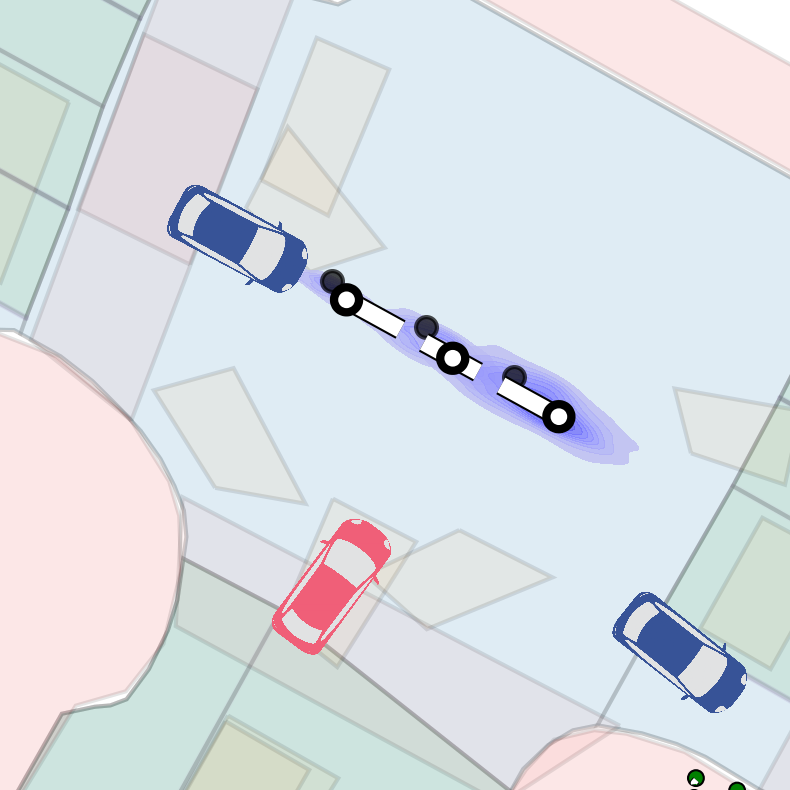}
    }
  \caption{\footnotesize Selected scenarios from nuScenes. The ego vehicle is red, non-ego vehicles are blue, pedestrians are green circles, ground truth trajectories are white, and Trajectron++ predictions are the purple contour plots. \label{fig:postive_detection_comparison}}
\end{figure}

\textbf{Results.} We compare our detector to all baseline methods described earlier. See Table~\ref{tab:filter-results} for a comparison of the methods and Fig.~\ref{fig:postive_detection_comparison} for representative examples of positive detections. 
We observe that the likelihood detection and the TTC-based detection both have many more positives than the other methods, followed by HJ reachability. Other methods have a similar number of positives. Ultimately, no driving scenario from this dataset is unsafe, so we inspect the positive detections from each method to find interesting scenarios. All methods detect a handful of interesting scenarios, such as Fig~\ref{fig:postive_detection_comparison}(b) where pedestrians are right beside the ego (possibly entering or exiting). We 
\begin{wraptable}{r}{0.45\textwidth}
\vspace{2mm}
\footnotesize
    \centering
    \caption{Detections from expert driving logs \label{tab:filter-results}}
    \begin{tabular}{cc}
        \toprule
        Detection Method & Percentage Detected\\
        \midrule
        UDT \citep{Greenberg21} & $5.4\pm 0.1$\\
        QAD (FPR) (ours) & $5.4\pm 0.1$\\
        PDT \citep{Greenberg21} & $5.8\pm 0.1$\\
        QAD (FNR) (ours) & $6.4\pm 0.1$ \\
        HJ Reachability \cite{LeungSchmerlingEtAl2019} & $8.8\pm0.0$\\
        TTC & $17.4\pm0.0$\\
        Likelihood & $58.5\pm0.0$\\
        \bottomrule
    \end{tabular} 
\end{wraptable}
also see additional interesting scenarios detected by our method but not HJ reachability such as in Fig.~\ref{fig:postive_detection_comparison}(a) where a close-by pedestrian is predicted to stand still, but moves quickly towards the ego at a crosswalk. Qualitatively, it seems that our detector, UDT, and PDT all do a good job of identifying interesting scenarios such as those shown in Fig.~\ref{fig:postive_detection_comparison}(a) and (b). We see that HJ reachability, and likelihood detection methods more frequently labels driving scenarios where the predictions are accurate and there is nothing inherently unsafe as anomalous such as in Fig.~\ref{fig:postive_detection_comparison}(c) and (d).  

%% file: sections/limitations-future-conclusion.tex
\section{Limitations, Future Work, and Conclusion}

\textbf{Limitations.} As discussed in Sec.~\ref{subsec:calibration}, we cannot provide a strong FPR and FNR guarantee simultaneously for our detector. This does not seem to be a fundamental limitation of the detector itself because our empirical results in Table~\ref{table:best_point_on_roc} suggest that we can achieve low FPR and low FNR. As part of our future work, we intend to explore providing stronger guarantees on both FPR and FNR simultaneously. A second limitation of our approach is identifying what quantile of the cost distribution corresponds to an actual anomaly. As we saw in Sec.~\ref{subsec:exp-task-relevant}, the mismatch in the quantile can result in a disagreement with the theoretical bounds. From a practical perspective, this is not a major limitation as we demonstrated that our detector performs well despite making a completely uninformed quantile choice for anomaly detection. Furthermore, we can mitigate this problem by using a small dataset to identify an appropriate $p$-quantile to be considered an anomaly. 

\textbf{Future Work.} As part of our future work, we hope to use this detector for adapting planning frequencies. If the planner re-plans only when necessary, it would free up some of the compute budget for other modules in the AV stack and lower the energy consumption. Since our detector directly monitors the cost of the planner, it would serve as a good indicator of when the motion plan is not evolving as expected and therefore, requires re-planning. As a proof of concept, we present some motivating results in App.~\ref{app:adaptive planning rate}. 
An additional direction for future work is to require multiple $p$-quantile anomalies before declaring a detection instead of a single one used in this work. Another exciting direction for this work is providing causal explanations for the detections; e.g., we anticipated the agent to continue straight, but it is switching into our lane. Such explanations can actively be leveraged to plan reactions to prediction failures. 

\textbf{Conclusion.} We presented a task-relevant detector for mispredictions by the trajectory forecasting networks used in AVs. Our detector builds a distribution of the predicted future costs and identifies an anomaly when the true cost of the planner lies in the top-p quantile of that distribution. We provided FPR and FNR guarantees for our detector that facilitate data-free calibration. In experiments, we demonstrated that our detector out-performs all other baseline approaches we compared against and is compute-efficient. Additionally, we showed that our data-free calibrated detector's performance is close to the optimal detector chosen from the ROC curve. We used our detector for run-time misprediction monitoring and for filtering prediction failures from unlabeled driving logs.

%% file: sections/appendix.tex
\appendix

\section{Proof of Theorem~\ref{thm:fpr-fnr}}
\label{app:proof}

\begin{proof}
We will first establish the bound on the FPR. Note that
\begin{align}
    c^\star_{t+\tau} \geq \hat{c}_{t+\tau}^{M-n} \wedge c^\star_{t+\tau} < c^{\rm a}_{t+\tau}(p) \implies \hat{c}_{t+\tau}^{M-n} < c^{\rm a}_{t+\tau}(p) \enspace.
\end{align}
Hence,
\begin{align}
    \underset{S\sim\phi^M}{\mathbb{P}}[c^\star_{t+\tau} \geq \hat{c}_{t+\tau}^{M-n} \wedge c^\star_{t+\tau} < c^{\rm a}_{t+\tau}(p)] \leq \underset{S\sim\phi^M}{\mathbb{P}}[\hat{c}_{t+\tau}^{M-n} < c^{\rm a}_{t+\tau}(p)] \enspace. \label{eq:fp-1}
\end{align}
Since $\hat{c}$'s are ordered by their value in $S$, the upper bound in \eqref{eq:fp-1} is essentially the probability of sampling $S\sim \phi_{t+\tau}(\hat{c}|o_t)^M$ such that at most $n$ elements of $S$ lie above $c^{\rm a}_{t+\tau}(p)$, i.e., lie in the $p$-quantile tail of $\phi$. Whether a sample of $\phi$ lies in the top $p$-quantile tail or the remainder of the $1-p$ portion of the distribution is Bernoulli distributed. Therefore,
\begin{align}
    \underset{S\sim\phi^M}{\mathbb{P}}[\hat{c}_{t+\tau}^{M-n} < c^{\rm a}_{t+\tau}(p)] = \sum_{i=0}^n \binom{M}{i} p^i (1-p)^{M - i} \enspace. \label{eq:fp-bernoulli}
\end{align}
Using \eqref{eq:fp-bernoulli} in \eqref{eq:fp-1} followed by \eqref{eq:fpr-def} gives the upper bound on FPR.

Analogously, we can bound the FNR. We have the following:
\begin{align}
    c^\star_{t+\tau} < \hat{c}_{t+\tau}^{M-n} \wedge c^\star_{t+\tau} \geq c^{\rm a}_{t+\tau}(p) \implies \hat{c}_{t+\tau}^{M-n} > c^{\rm a}_{t+\tau}(p) \implies \hat{c}_{t+\tau}^{M-n} \geq c^{\rm a}_{t+\tau}(p) \enspace.
\end{align}
Hence,
\begin{align}
    \underset{S\sim\phi^M}{\mathbb{P}}[c^\star_{t+\tau} < \hat{c}_{t+\tau}^{M-n} \wedge c^\star_{t+\tau} \geq c^{\rm a}_{t+\tau}(p)] \leq \underset{S\sim\phi^M}{\mathbb{P}}[\hat{c}_{t+\tau}^{M-n} \geq c^{\rm a}_{t+\tau}(p)] \enspace. \label{eq:fn-1}
\end{align}
As before, we use the fact that drawing samples from $\phi$ such that they lie in the $p$ or the $1-p$ portions of the distribution is Bernoulli distributed to write
\begin{align}
    \underset{S\sim\phi^M}{\mathbb{P}}[\hat{c}_{t+\tau}^{M-n} \geq c^{\rm a}_{t+\tau}(p)] = \sum_{i=n+1}^M \binom{M}{i} p^i (1-p)^{M - i} \enspace. \label{eq:fn-bernoulli}
\end{align}
Using \eqref{eq:fn-bernoulli} in \eqref{eq:fn-1} followed by \eqref{eq:fnr-def} gives the upper bound on FNR, completing the proof.
\end{proof}

\section{Additional Experimental Details}

\subsection{Comparison methods}
\label{app:comparisons}
We compare our detector against four other approaches: (i) likelihood detection, (ii) uniform and partial degradation tests (UDT and PDT) \citep{Greenberg21}, (iii) a time-to-collision (TTC) based detector, and (iv) detection based on Hamilton-Jacobi (HJ) reachability analysis \cite{LeungSchmerlingEtAl2019}.

\textbf{Likelihood detection.} We label the likelihood of the estimated state $x_{t+\tau}$ as the value in the PDF of the distribution over the predicted non-ego agent states $\psi(\hat{x}_{t+1:t+T}^\text{ne}|x_{0:t})$ provided by the prediction module. As such, we directly represent the likelihood of the estimated non-ego positions given the predicted positions. For anomaly detection, we only check agents within $10$ m of the ego since this is the region in which trajectory prediction is most accurate, and the non-ego agents are most relevant.

\textbf{UDT and PDT.} These degradation tests use the predicted costs for each agent as a set of reference signals. We use the same cost function as our method. UDT compares a weighted mean of the reference signals to the online signal in order to determine when reward degradation is significant and is optimal when the reference signals come from a multivariate normal distribution. PDT detects deterioration using a subset of the time steps of the reference signal and tends to perform better in practice when the normality assumption does not hold, see \cite{Greenberg21} for more information on these tests. After predicted costs are computed, we use the false alarm rate control method provided in \citep{Greenberg21} to choose a threshold for detection based on the desired $p$-value and the reference (i.e. predicted) costs. If the achieved cost is above the threshold, it is labeled as an anomaly.

\textbf{TTC detection.} For TTC calculations, we assume vehicles are a $1$m radius circle and pedestrians are a $0.2$m radius circle. We propagate the agents forward in time to determine if there would be a collision at some point in the future if the agents maintained their velocity along the current heading.

\textbf{HJ reachability.} We follow the method for HJ reachability analysis from \cite{LeungSchmerlingEtAl2019}. We assume the ego and other vehicles follow a four-state bicycle model with the following dynamics
\begin{align}
    [\dot{x},\dot{y}, \dot{\theta}, \dot{v}]  = [v \cos(\theta), v \sin(\theta), \tan(d) / L, a]
\end{align}
Where $d$ is the control steering angle, $a$ is the control acceleration, and $L$ is the length of the vehicle. We assume that the vehicles' acceleration lies within $[-2, 1] \text{ ms}^{-2}$ and the steering angle lies within $[-10,10]$ degrees. Pedestrians are modeled as agents which can move in any direction with acceleration in $[-0.5, 0.5] \text{ ms}^{-2}$. We assume a $1$m radius circle for vehicles and a $0.2$m radius circle for pedestrians and set the initial value function such that any collision should result in a negative value. We compute the value function offline to improve operation speed. Online, we check the value for each agent with the ego using a lookup table. If the lowest value is below a threshold, we label the scene as anomalous. 

\subsection{Planning}
\label{app:planning}
We follow a similar procedure as in \citep{SchmerlingLeungEtAl2018} to make a motion plan for the ego. First, we discretize the control space so that we can generate a tree of motion primitive trajectories by integrating along all possible combinations of discrete control actions at each time step over the planning horizon. In particular, we use $[-2, 0, 1] \text{ ms}^{-2}$ for the acceleration control and $[-0.3, -0.1, 0, 0.1, 0.3] \text{ rad s}^{-1}$ for the radial velocity control as the discretization. To allow for time to run cost computations, we set the first control action of each primitive to be the most recent control action as in \citep{SchmerlingLeungEtAl2018}. As such, there are $15^{(T-1)}$ possible motion primitives where $T$ is the planning horizon. We compute the predicted cost for each motion primitive using predictions of all other agents in the scene and select the primitive which has the lowest predicted cost over the planning horizon. We parallelize the predicted cost computations to improve run-time. A cost computation for a single primitive takes about $1\pm 0.5$ms and for a planning horizon of 2 seconds (4 control inputs), there are $3375$ motion primitives ($15^{3}$) which need cost computations. On the 18 core (36 thread) computer we use for simulations, with parallelization, the best motion primitive can be reliably computed in less than $0.25$s, which allows for real-time planning.

\subsection{Cost functions}
\label{app:cost-func}
Let $x_e, v_e, a_e, j_e, \theta_e, \dot\theta_e, \ddot\theta_a$ be the position, velocity, acceleration, jerk, heading, rotational velocity, and rotational acceleration of the ego, respectively. Similarly, $x_a, v_a, a_a, j_a, \theta_a, \dot\theta_a,\ddot\theta_a$ represent the analogous states for the non-ego agents. Note that when we compute predictions of the cost, we use the predicted non-ego agent states instead of the achieved agent states in the cost computation; otherwise, the cost computations are identical for the predicted and the observed costs. We add subscript $\parallel$ or $\perp$ to denote decomposition parallel to or perpendicular to the \textit{ego's heading} (e.g. ego's longitudinal velocity $v_{e,\parallel}$ and agent's lateral acceleration with respect to the ego's heading $a_{a,\perp}$). We define $x_\text{goal}$ as the position of the end of the reference trajectory (i.e., the goal), and $x_r, \theta_r$ are points and headings on the reference trajectory. Let $\epsilon$'s represent scaling factors and let $\text{ttc}$ represent the time-to-collision function; ttc outputs the time until a collision between the ego and another agent (assuming constant velocity at their current heading direction) and modeling vehicles as a $1$ m radius circle and pedestrians as a $0.2$ m radius circle. We now express the cost functions we use as follows:

{\allowdisplaybreaks
\begin{align}
    c_{\text{ttc}} =& 1 - \max_{\text{non-ego agent}}\min\Big(\frac{\text{ttc}(x_e, x_a, v_e, v_a)}{\epsilon_{\text{ttc}}}, 1\Big)  \enspace, \\ 
    c_{\text{d2a}} =& \max_{\text{non-ego agent}}\Big[ e^{-0.5 \epsilon_\text{rbf} ((x_{a,\parallel} - x_{e, \parallel})^2(v_{a,\parallel} - v_{e, \parallel})^2  +  (x_{a,\perp} - x_{e,\perp})^2(v_{a,\perp} - v_{e,\perp})^2)} \Big]  \enspace, \\ 
    c_{\text{d2g}} =& \|x_\text{goal} - x_e\| /\|x_\text{goal} - x_{0,e}\| \text{ where } x_{0,e} = x_e \text{ at } t=0 \enspace, \\ 
    c_{\text{d2r}} = &  \frac{1}{4}\|x^*_r - x_e\|^4 + \frac{1}{2}(\theta^*_r - \theta_e)^2 \text{ where } x^*_r, \theta^*_r= \underset{(x_r, \theta_r) \in \text{reference trajectory}}{\text{argmin}} \|x_r - x_e\| \enspace \\
    c_{\text{velocity}} =& \max(\|v_e - \epsilon_{v,r}\| - \epsilon_r, 0)^2 / \epsilon_{\text{limit}}^2 \enspace, \\
    c_{\text{comfort}} =&\frac{1}{6}\sum \max\Bigg(\bigg[
    \frac{|a_{e,\parallel}|}{\epsilon_{a,\parallel}}-1,
    \frac{|a_{e,\perp}|}{\epsilon_{a,\perp}}-1, \frac{|j_{e,\parallel}|}{\epsilon_{j,\parallel}}-1,
    \frac{\|j_e\|}{\epsilon_j}-1,
    \frac{|\dot{\theta}_e|}{\epsilon_{\dot\theta}}-1,
    \frac{|\ddot{\theta}_e|}{\epsilon_{\ddot\theta}}-1\bigg],
    \bar{0}\Bigg) \label{eq:costfn comfort} \enspace, \\ 
    c_{\text{reverse}} =& 1[v_{e,\parallel} < 0] \enspace. 
\end{align}}

In our experiments, we use $\epsilon_\text{ttc} = 3$. $\epsilon_\text{rbf} = 0.5$ when the agent is a vehicle, and $\epsilon_\text{rbf} = 1$ when the agent is a pedestrian. The reference velocity $\epsilon_{v,r} = 0.8 \|x_\text{goal} - x_{0,e}\| / \text{time of reference trajectory}$, and $\epsilon_r = \max(0.1\epsilon_{v,r},1)$, $\epsilon_\text{limit} = \max(30 - \epsilon_{v,r}, 10)$. The comfort tuning parameters are based on the built-in comfort metric from nuPlan \citep{nuplan}, $(\epsilon_{a,\parallel},\epsilon_{a,\perp},\epsilon_{j,\parallel},\epsilon_{j},\epsilon_{\dot\theta},\epsilon_{\ddot\theta}) = (2.4, 4.89, 4.13, 8.37, 0.95, 1.93)$ and the $\max$ in \eqref{eq:costfn comfort} refers to element-wise maximum. Along with the tuning parameters, the weights are set as $[w_1, \dots, w_7] = [1,10,1,1,1,0.5,10]$.

\subsection{Hand labeling scenarios}
\label{app:hand_label}
\begin{figure}[h]
  \centering
    \subfigure[]{
    \includegraphics[width=0.30\textwidth]{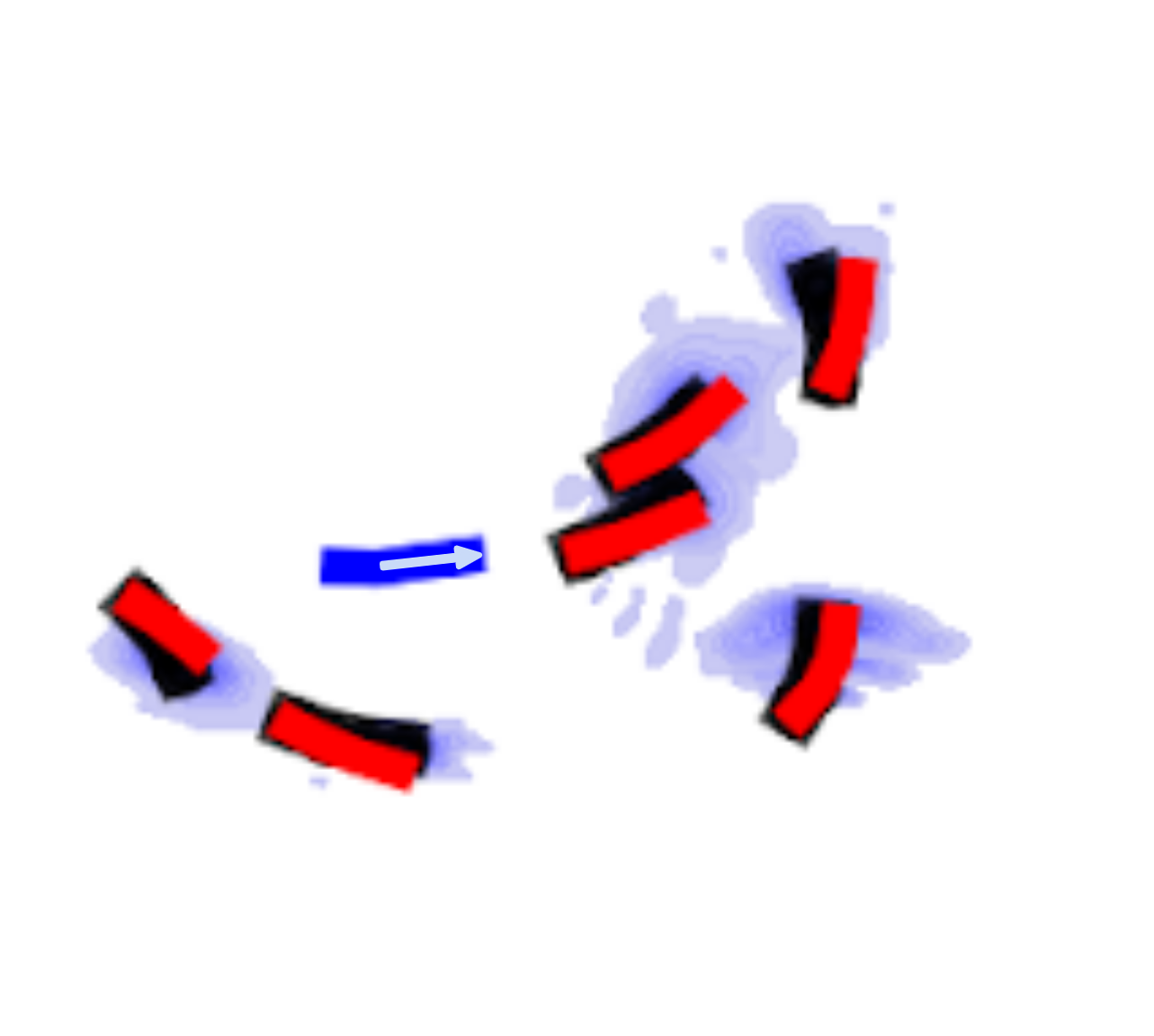}
    }    
    \subfigure[]{
    \includegraphics[width=0.30\textwidth]{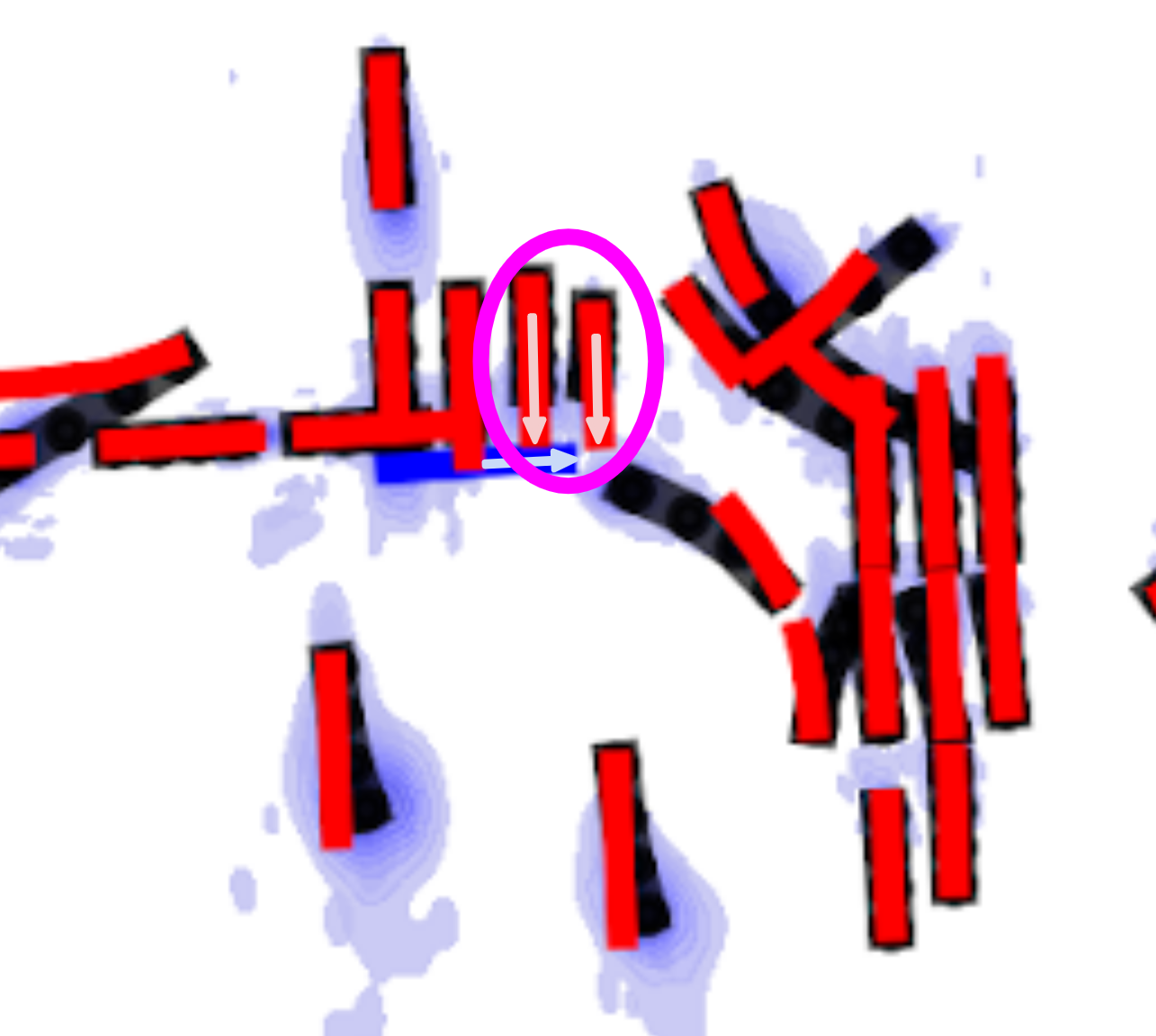}
    }
    \subfigure[]{
    \includegraphics[width=0.30\textwidth]{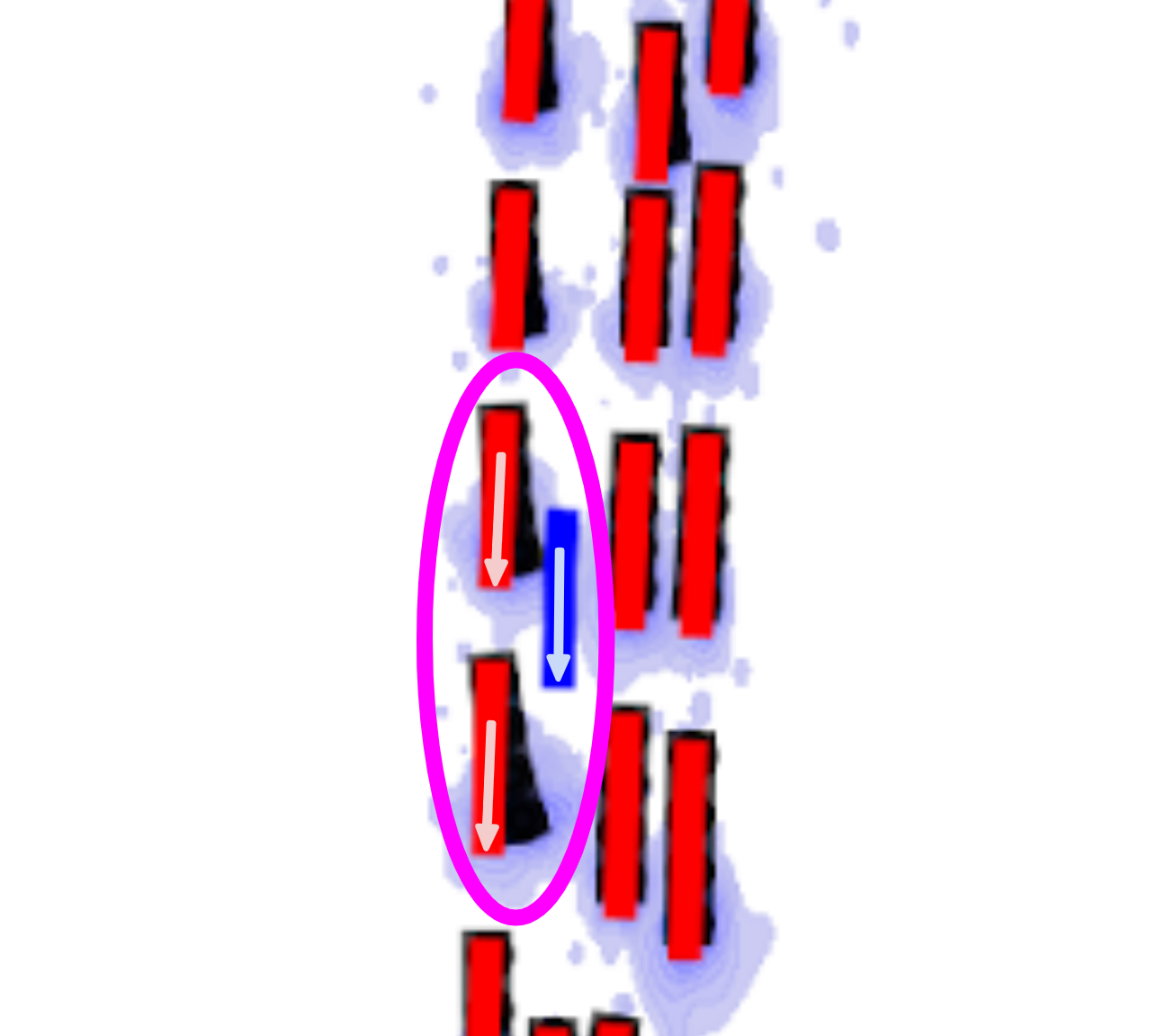}
    }
  \caption{\footnotesize Illustration of hand labeling through selected scenarios from nuPlan. The ego vehicle trajectory is in blue, non-ego vehicle trajectories are in red, arrows indicate direction of travel, Trajectron++ predictions are the purple contour, and the black line is the most likely trajectory given the predictions. Important agents are circled. (a) Example of a negative label. (b) Example of a positive label. (c) Example of a negative label with task-irrelevant prediction errors. \label{fig:labeling_scenarios}}
  \vspace{5pt}
\end{figure}

See Fig.~\ref{fig:labeling_scenarios} for examples of positive and negative hand labels. 
Fig.~\ref{fig:labeling_scenarios}(a) is an example of a negative label since agents are not close to the ego and predictions are relatively accurate. Fig.~\ref{fig:labeling_scenarios}(b) is an example of a positive label. The circled agents are predicted to slow down or stop before reaching the ego. Instead, these agents get very close to the ego. Lastly, Fig.~\ref{fig:labeling_scenarios}(c) is an example of a negative label with task-irrelevant prediction errors. Non-ego agents which are circled are predicted to turn into the ego's lane, but they do not. 

\subsection{Additional experimental results: Representative success and failure cases}
\begin{figure}[h]
  \centering
    \subfigure[True negatives (TN)]{
    \includegraphics[width=0.24\textwidth]{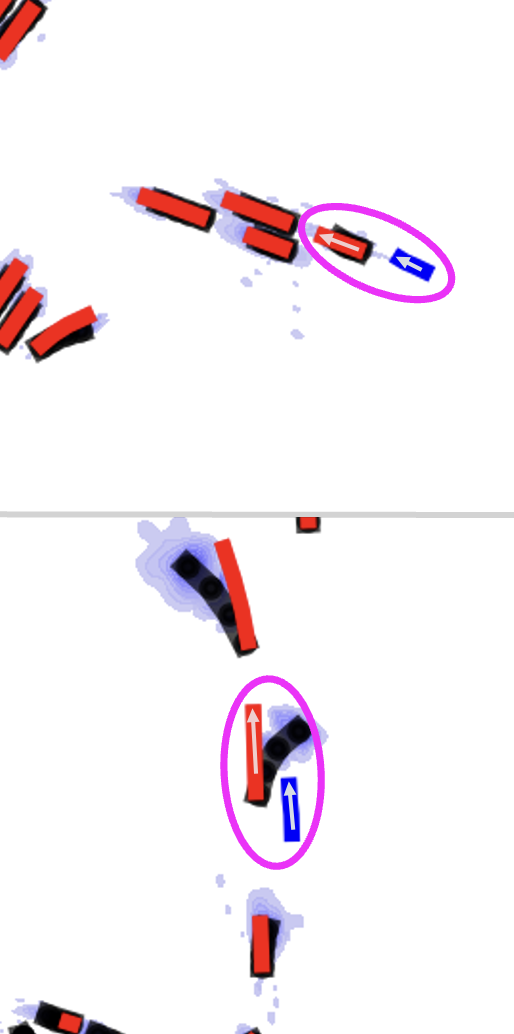}
    }\hspace{-0.5em}
    \subfigure[True positives (TP)]{
    \includegraphics[width=0.24\textwidth]{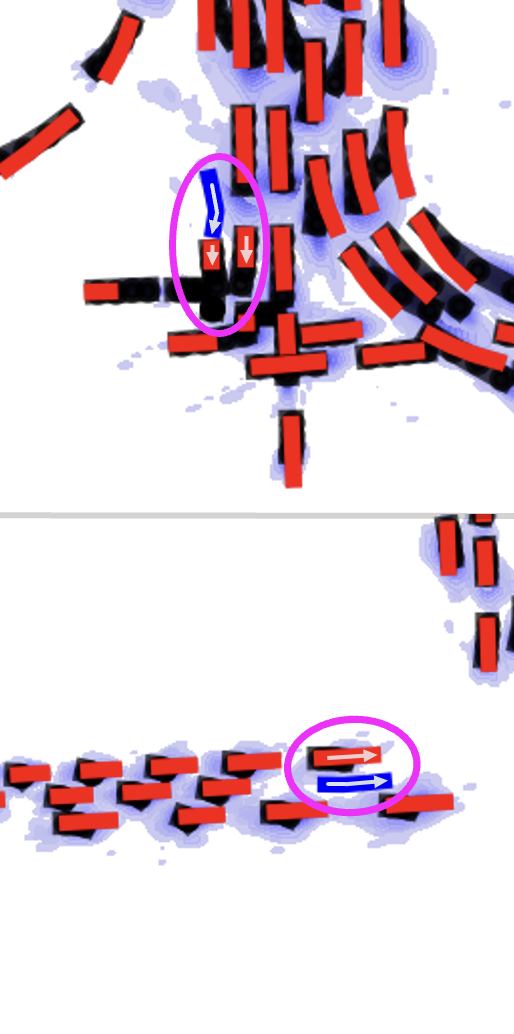}
    }\hspace{-0.5em}
    \subfigure[False negatives (FN)]{
    \includegraphics[width=0.24\textwidth]{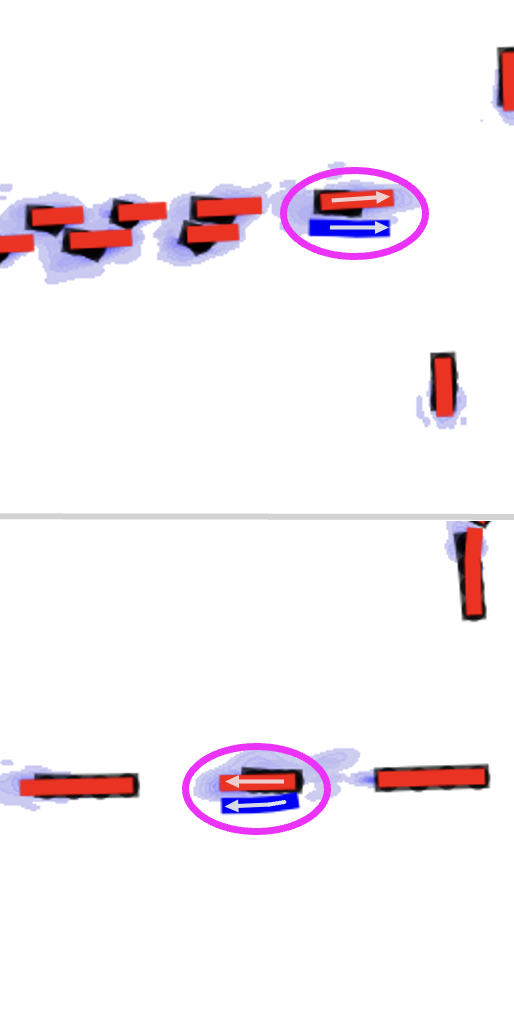}
    }\hspace{-0.5em}
    \subfigure[False positives (FP)]{
    \includegraphics[width=0.24\textwidth]{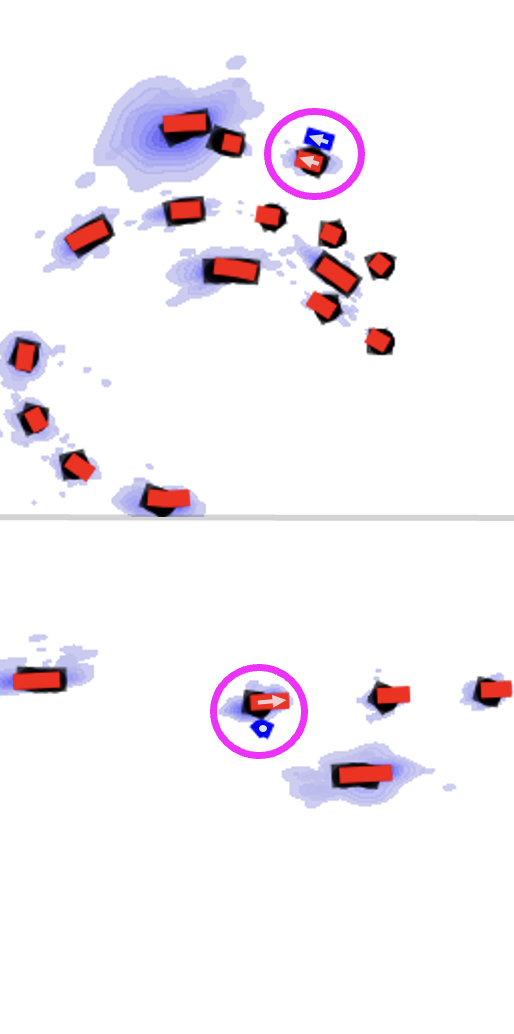}
    }
  \caption{\footnotesize Illustration of representative (a) true negatives (TP), (b) true positives (TP), (c) false negatives (FN), and (d) false positives (FP) from the nuPlan dataset. The ego vehicle trajectory is in blue, non-ego vehicle trajectories are in red, arrows indicate direction of travel, Trajectron++ predictions are the purple contour, and the black line is the most likely trajectory given the predictions. Important agents are circled.
  \label{fig:tntpfnfp}}
  \vspace{5pt}
\end{figure}

See Fig.~\ref{fig:tntpfnfp} for representative examples of true negatives, true positives, false negatives, and false positives. The images provide qualitative information about where the failure predictor succeeds and fails. The true negatives in Fig.~\ref{fig:tntpfnfp}(a) show standard driving scenarios and the ability of our detector to ignore task-irrelevant prediction errors. The true positives in Fig.~\ref{fig:tntpfnfp}(b) are scenarios where other agents are predicted to maintain their speed but speed up or slow down more quickly. The false negatives in Fig.~\ref{fig:tntpfnfp}(c) are cases where predictions have are incorrect but have larger variance. The images are labeled as positive by the hand-labeling, but the higher variance is enough to prevent a $p$-quantile anomaly detection from occurring. The false positives in Fig.~\ref{fig:tntpfnfp}(d) are cases which look like normal driving scenarios, but anomalies are detected due to the very close agents and small variance in predictions.

\subsection{Additional experimental results: Adaptive re-planning rate}
\label{app:adaptive planning rate}
\begin{figure}[h]
\vspace{-10pt}
  \centering
    \includegraphics[width=0.65\textwidth]{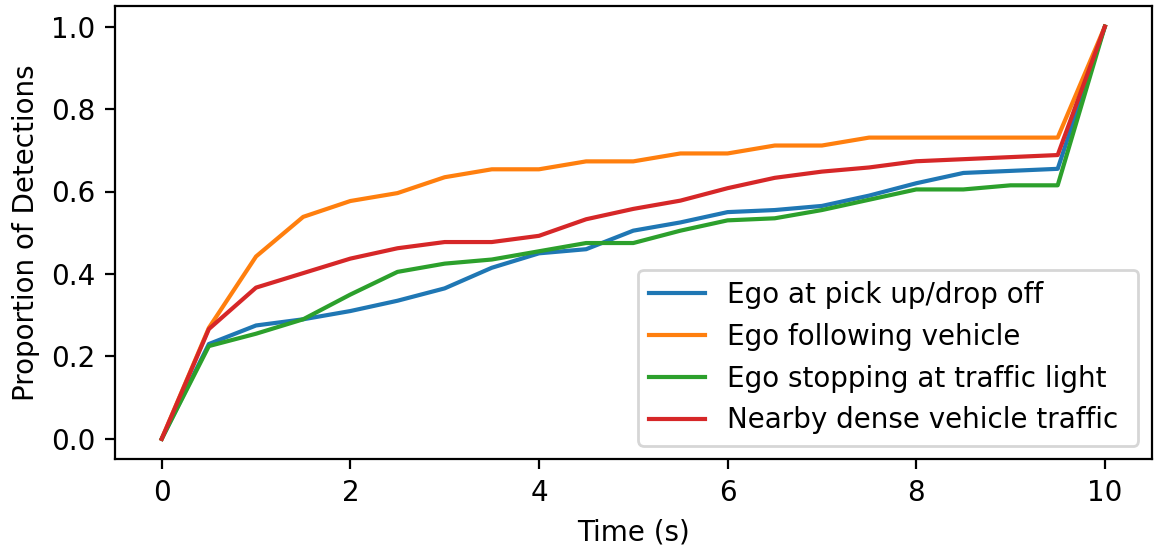}
  \caption{\footnotesize Cumulative distribution function for the length of time between re-planning for various types of scenarios. For ego following vehicle scenes (orange line), there is more frequent re-planning. For the ego stopping at traffic light (green line) and ego at pick up/drop off (blue line) scene types, re-planning is less frequent. \label{fig:adaptive_replanning}}
\end{figure}

In order to assess if our detector can feasibly be used for an adaptive re-planning rate, we set up another experiment. Initially, the ego is given a long horizon plan (10 seconds, 20 time steps) from the reference trajectory. We run our quantile anomaly detector with quantile $p=0.25$ and re-plan once an anomaly is detected. The result is a detector which determines when the predictions are incorrect enough such that re-planning may be required. The dataset is made of 651 scenarios which come from 4 classes of scenario types: (i) ego at pick-up/drop-off (PUDO), (ii) ego following vehicle, (iii) ego stopping at traffic light, and (iv) nearby dense vehicle traffic. See Fig.~\ref{fig:adaptive_replanning} for a cumulative distribution function (CDF) of detections for each scenario type. We also present the mean detection time for each of the scene types in Table~\ref{tab:adaptive_planning}. We see that the ego following vehicle scene type has the lowest time between re-plans while the ego stopping at traffic light and ego at pick up/drop off both scene types have, on average, longer time between re-planning. The nearby dense traffic scenarios tend to have many more vehicles and the ego following vehicle has many more lane-changes and a large diversity of agent speeds. This results in a larger number of task-relevant prediction failures and we therefore see faster re-planning rate for these scenarios. In contrast, we observe that the ego at PUDO and the ego stopping at traffic light scenarios have many fewer non-ego agents around as compared with the other scenario types and therefore have fewer task-relevant prediction failures. The long average time duration between re-plans in Table~\ref{tab:adaptive_planning} suggest the viability of our approach as an effective way to adapt the re-planning rate.

\begin{table}[h]
\vspace{-5pt}
    \centering
    \caption{Average time between re-plans \label{tab:adaptive_planning}}
    \begin{tabular}{ccccc}
        \toprule
        Scene type: & Ego at PUDO & Following vehicle & Stopping at light & Nearby traffic \\
        \midrule
        Mean re-plan time & 5.18 s & 3.55 s & 5.23 s & 4.49 s \\
        \bottomrule
    \end{tabular}
\end{table}

